\documentclass[10pt,twocolumn,letterpaper]{article}

\usepackage{cvpr}
\usepackage{times}
\usepackage{epsfig}
\usepackage{graphicx}
\usepackage{amsmath}
\usepackage{amssymb}
\usepackage{makecell}

\usepackage{adjustbox}

\usepackage{enumitem,kantlipsum}
\usepackage{multicol}
\usepackage{multirow}
\usepackage{caption}
\usepackage{subcaption}
\usepackage{array}
\usepackage{booktabs}
\usepackage{color}
\usepackage{kotex}
\usepackage{soul}
\usepackage{setspace}
\usepackage{algorithm}
\usepackage{algorithmic}

\usepackage{bm}
\usepackage{comment}

\usepackage{mathtools}

\renewcommand{\algorithmiccomment}[1]{\bgroup\hfill//~#1\egroup}

\usepackage[pagebackref=true,breaklinks=true,letterpaper=true,colorlinks,bookmarks=false]{hyperref}

\cvprfinalcopy % *** Uncomment this line for the final submission

\def\cvprPaperID{7592} % *** Enter the CVPR Paper ID here

\newcommand\nj[1]{\textcolor{red}{#1}}
\newcommand\jh[1]{\textcolor{cyan}{#1}}
\newcommand\yb[1]{\textcolor{green}{#1}}

\ifcvprfinal\fi
\begin{document}

%%%%%%%%%%%%%%%%%   symble

%%%%%%%%%%%%%%%%5

%%%%%%%%% TITLE
\title{QKD: Quantization-aware Knowledge Distillation% for Quantized Networks
}

% \author{Jangho Kim\thanks{Equal contribution} \thinspace \thanks{Work done while interning at Qualcomm AI Research.}\\
% Seoul National Universitiy
% \\
% Seoul, Korea\\
% {\tt\small kjh91@snu.ac.kr}
% % For a paper whose authors are all at the same institution,
% % omit the following lines up until the closing ``}''.
% % Additional authors and addresses can be added with ``\and'',
% % just like the second author.
% % To save space, use either the email address or home page, not both
% \and
% Yash Bhalgat\footnotemark[1]\\
% Qualcomm AI Research\\
% San Diego, USA\\
% {\tt\small ybhalgat@qti.qualcomm.com}
% \and
% Jinwon Lee\\
% Qualcomm AI Research\\
% San Diego, USA\\
% {\tt\small jinwonl@qti.qualcomm.com}
% \and
% Chirag Pate\\
% Qualcomm AI Research\\
% San Diego, USA\\
% {\tt\small cpatel@qti.qualcomm.com}
% \and
% Nojun Kwak\\
% Seoul National Universitiy
% \\
% Seoul, Korea\\
% {\tt\small nojunk@snu.ac.kr}
% }

\author{Jangho Kim\thanks{Equal contribution} \thinspace \thanks{Work done while interning at Qualcomm AI Research.}\\
Seoul National University
\\
{\tt\small kjh91@snu.ac.kr}
% For a paper whose authors are all at the same institution,
% omit the following lines up until the closing ``}''.
% Additional authors and addresses can be added with ``\and'',
% just like the second author.
% To save space, use either the email address or home page, not both
\and
Yash Bhalgat\footnotemark[1]\\
Qualcomm AI Research\thanks{\scriptsize Qualcomm AI Research is an initiative of Qualcomm Technologies, Inc.}\\
Qualcomm Technologies, Inc.\\
{\tt\small ybhalgat@qti.qualcomm.com}
\and
Jinwon Lee\\
Qualcomm AI Research\footnotemark[3]\\
Qualcomm Technologies, Inc.\\
{\tt\small jinwonl@qti.qualcomm.com}
\and
Chirag Patel\\
Qualcomm AI Research\footnotemark[3]\\
Qualcomm Technologies, Inc.\\
{\tt\small cpatel@qti.qualcomm.com}
\and
Nojun Kwak\thanks{Currently a Visiting Researcher at Qualcomm Technologies, Inc.}\\
Seoul National University \\
{\tt\small nojunk@snu.ac.kr}
}

\maketitle
%\thispagestyle{empty}

%%%%%%%%% ABSTRACT
\begin{abstract}
\begin{comment}
Deep neural networks (DNNs) have achieved significant advances in many domains. %such as computer vision and natural language processing. 
However,  large model sizes and huge computational costs make it challenging to deploy DNNs on resource-limited edge devices.
\end{comment}
%are
%\nj{related to power consumption} 
%still \nj{problems} \nj{of DNN},
%which should be reduced for resource\nj{-limited devices} such as mobile phone\nj{s}.

% To reduce memory and power consumption of deep neural networks (DNNs), model compression, quantization, or knowledge distillation (KD) methods or their combinations are widely used. However, combining quantization and distillation may not work as desired %can cause underfitting 

% We propose quantization-aware knowledge distillation (QKD). The key ideas of QKD is in three-folds. The first is Self-Studying (SS) wherein we trained the student without KD to obtain good initialization. The second is Co-Studying (CS) wherein we even trained teacher to make it more quantizaton-friendly. Finally, Tutoring (TU) transfer knowledge from the trained teacher to the student. By combining SS, CS, TU will improves accuracy a lot compared to conventional KD for quantization.

Quantization and Knowledge distillation (KD) methods are widely used to reduce memory and power consumption of deep neural networks (DNNs), especially for resource-constrained edge devices. Although their combination is quite promising to meet these requirements,
%further compress the model,
it may not work as desired. It is mainly because the regularization effect of KD further diminishes the already reduced representation power of a quantized model. To address this shortcoming, we propose Quantization-aware Knowledge Distillation (QKD) wherein quantization and KD are carefully coordinated in three phases. First, Self-studying (SS) phase fine-tunes a quantized low-precision student network without KD to obtain a good initialization. Second, Co-studying (CS) phase tries to train a teacher to make it more quantizaion-friendly and powerful than a fixed teacher. Finally, Tutoring (TU) phase transfers knowledge from the trained teacher to the student. We extensively evaluate our method on ImageNet and CIFAR-10/100 datasets and show an ablation study on networks with both standard and depthwise-separable convolutions. The proposed QKD outperformed existing state-of-the-art methods (e.g., 1.3\% improvement on ResNet-18 with W4A4, 2.6\% on MobileNetV2 with W4A4). Additionally, QKD could recover the full-precision accuracy at as low as W3A3 quantization on ResNet and W6A6 quantization on MobilenetV2. (See Fig. \ref{fig:first})
\end{abstract}

\section{Introduction}
\label{introduction}
\begin{comment}
Deep neural networks (DNNs) are widely used across several domains such as computer vision, natural language processing today.
\end{comment}

Deploying complex DNNs on resource-constrained edge devices such as smartphones or IoT devices is still challenging due to their tight memory and computation requirements. To address this, several research works have been proposed which can be roughly categorized into three folds: weight pruning \cite{han2015learning,han2015deep,liu2018rethinking,frankle2018lottery}, quantization, %which reduces bit-widths of weights and activations
\cite{krishnamoorthi2018quantizing,zhang2018lq,lin2016fixed,wang2018two,cai2017deep} and knowledge distillation %, which transfers knowledge from a high complexity teacher network to a low complexity student network
\cite{hinton2015distilling,kim2018paraphrasing,yim2017gift}.

\begin{figure}
  \begin{subfigure}[b]{0.23\textwidth}
    \includegraphics[width=\textwidth]{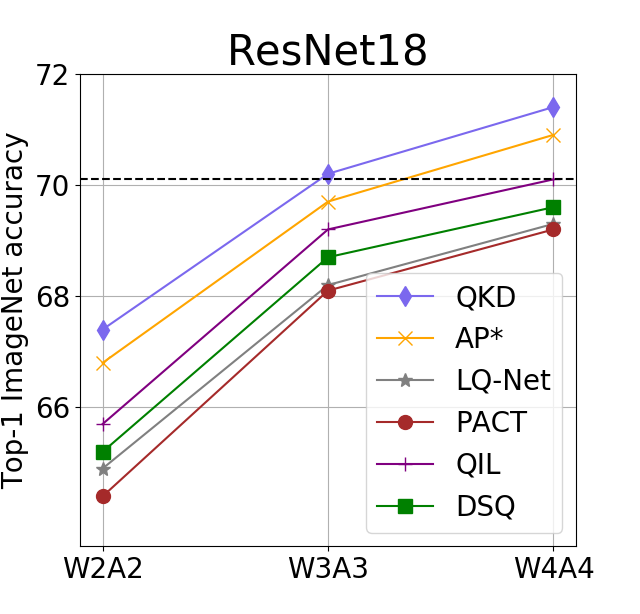}
    \label{fig:1}
  \end{subfigure}
  \begin{subfigure}[b]{0.23\textwidth}
    \includegraphics[width=\textwidth]{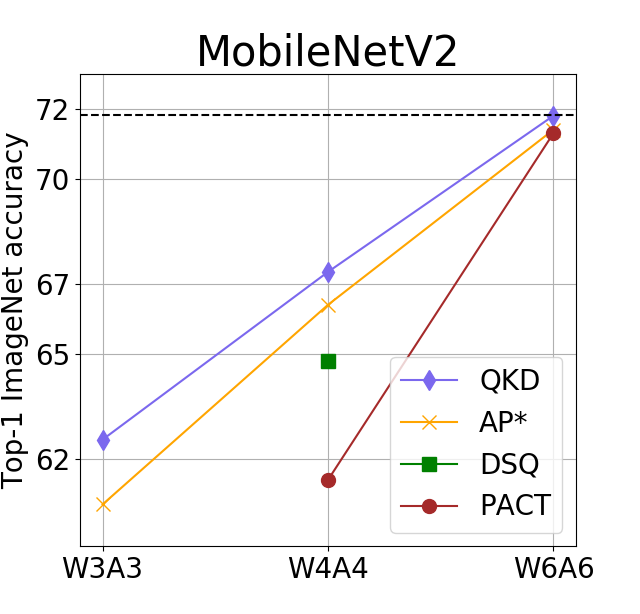}
    \label{fig:2}
  \end{subfigure}
  \caption{Top-1 accuracy on ImageNet with ResNet-18 and MobileNetV2. Our QKD is compared with various methods such as AP* \cite{mishra2017apprentice}, LQ-Net \cite{zhang2018lq}, PACT \cite{choi2018pact}, QIL \cite{jung2019learning} and DSQ \cite{gong2019differentiable}. Dotted line indicates the full-Precision accuracy. More details are in Table \ref{table:standard} and \ref{table:depthwise}.}
  \label{fig:first}
\end{figure}

% \begin{table}[t]
% 	\small
% 	\centering
% 	\begin{adjustbox}{width=0.6\linewidth}
% 	\begin{tabular}{p{1.1cm} c c c c}
%         \toprule
%         	{Method} & {W2A2} & {W3A3} & {W4A4} & W6A6\\
			
% 			\midrule
	
% 			QKD & 45.7 & 62.6 & 67.3 & 71.8 \\
% 			BL & 37.7 & 60.1 & 66.1 & 71.2  \\
% 			DSQ \cite{gong2019differentiable} & - & - & 64.8 & -  \\
% 			PACT \cite{choi2018pact} & - & - & 61.4 & 71.3 \\
%         \bottomrule
% 	\end{tabular}
% 	\end{adjustbox}
% 	\caption{MobileNetV2 ImageNet top-1 accuracy}
% 	\label{table:MV2_acc}
% \end{table}

Today, a majority of edge devices require fixed-point inference for compute- and power-efficiency. Hence, quantizing the weights and activations of deep networks to a certain bit-width becomes a necessity to make them compatible to edge devices.
%\nj{Among them, quantization is o}ne of key technique\nj{s that enables a} model to run in \nj{an} edge device. % is quantization.\nj{Until recently, most edge devices have only supported fixed-point operations which result in higher power consumption and longer latency. } %DSP or NPU which only support fixed point models, and much provide much lower power processing and low latency processing. 
Recently, hardware accelerators like NPUs (Neural Processing Units), NVIDIA's Tensor Core units and even CIM (Compute-in-memory) devices have targeted support for sub-byte level processing such as 4-bit$\times$4-bit or 2-bit$\times$2-bit matrix multiply-and-accumulate operations \cite{markidis2018nvidia,pan2018multilevel,sharma2019accelerated,sumbul2019compute}. These are much more power- and compute-efficient than conventional 8-bit processing. Thus, there are increased demands for lower-bit quantization.

\begin{comment}
Quantization reduces memory required for storing model parameters by replacing float-point values with low-bit representations. Further, quantization improves on-target inference latency and power consumption because typical HW accelerators (e.g., DSP, Neural Processing Units (NPUs)) are more efficient with low-bit representation multiply-accumulate (MAC) operations.
\end{comment}

\yb{}

% Although quantization reduces memory footprint, inference latency, and power comsumption,

% improves on-target inference latency as well as power consumption because typical HW accelerators (e.g., DSP and NPU) are more efficient with low-bit multiply-accumulate (MAC) operations, 
% %However, 
The accuracy of very low-bit quantized networks such as using 4-, 3- or 2-bits inevitably decreases because of their low representation power compared to typical 8-bit quantization. Some works \cite{mishra2017apprentice,polino2018model} used knowledge distillation (KD) to improve the accuracy of quantized networks by transferring knowledge from a full-precision teacher network to a quantized student network. However, even though KD is one of the widely used techniques to enhance a DNN's accuracy \cite{hinton2015distilling}, directly applying KD to very low-bit quantized networks incurs several challenges: 
%is not well-suited due to KD's effect as a regularizer.
 % in applying KD on top of quantized networks:
%
\begin{enumerate}[leftmargin=*]
    \itemsep0pt
    \item Due to its limited representation power, quantized networks generally show lower training and test accuracies compared to the networks with full precision. On the other hand, because KD uses not only the ground truth labels but also the teacher's estimated class probability distributions, it acts as a heavy regularizer \cite{hinton2015distilling}. Combining these contradictory characteristics of quantization and KD could sometimes result in further degradation of performance %compared to just vanilla quantization.
    from that of quantization-only.

    % \item Because of the inherent difference between  full-precision teacher network and quantized low-precision student network, their distributions of weights and activations are significantly different each other. Accordingly, it is hard to transfer the teacher's high-quality feature map (i.e., activation) to the student as pointed in \cite{mishra2017apprentice,polino2018model}.

    \item In general, it has been shown that a powerful teacher with high accuracy can teach a student better than a weaker one \cite{zhu2018knowledge}. Also, \cite{kim2018paraphrasing,mirzadeh2019improved} show that if there is a large gap from the capacity or inherent differences between the teacher and student, it can hinder the knowledge transfer because this knowledge is unadaptable to the student. Previous works of quantization with KD \cite{mishra2017apprentice,polino2018model} directly applied conventional KD to quantization along with fixed teacher network, and thus they suffer from above mentioned limitations.
    
    %use a fixed teacher which has these limitations in terms of powerful and adaptable knowledge that can be transferred for training a low-bit student network. %They do not consider above two characteristic\nj{s} of KD.
    
    %Previous works of quantization with KD \cite{mishra2017apprentice,polino2018model} use a fixed teacher which has these limitations in terms of powerful and adaptable knowledge that can be transferred for training a low-bit student network. %They do not consider above two characteristic\nj{s} of KD.

    %the teacher's high quality dark knowledge could not be transferred to the student as pointed in   , has limitation in transferring high-quality information fitted to the student.It is hard to apply feature\nj{-}map (activations) distillation  \cite{kim2018paraphrasing,romero2014fitnets} during training of a low-precision network because the distribution\nj{s} of weights and activations are significantly different compared to \nj{those of} a full-precision network (See Section \ref{discussion}}).
    
%    \item It is hard to apply feature\nj{-}map (activations) distillation  \cite{kim2018paraphrasing,romero2014fitnets} during training of a low-precision network because the distribution\nj{s} of weights and activations are significantly different compared to \nj{those of} a full-precision network (See Section \ref{discussion}}).

\end{enumerate}

\begin{comment}
Due to its limited representation power, the training accuracy of a low-bit DNN is quite low compared to its full-precision counterpart. Therefore, applying KD which acts as a strong regularizer to a low-bit DNN can cause more performance drop %more severe underfitting 
\jh{because of its limited representation power and the regularization effect of KD.}
% \nj{B}ecause \nj{a l}ow-bit DNN lose\nj{s} its representation power \nj{making} its training accuracy quite lower than full-precision DNN and KD acts as a hard regulizer in DNN, which easily can cause underfitting.
\end{comment}

 \begin{figure*}[t]
  \centering
  \includegraphics[width = 0.85\linewidth]{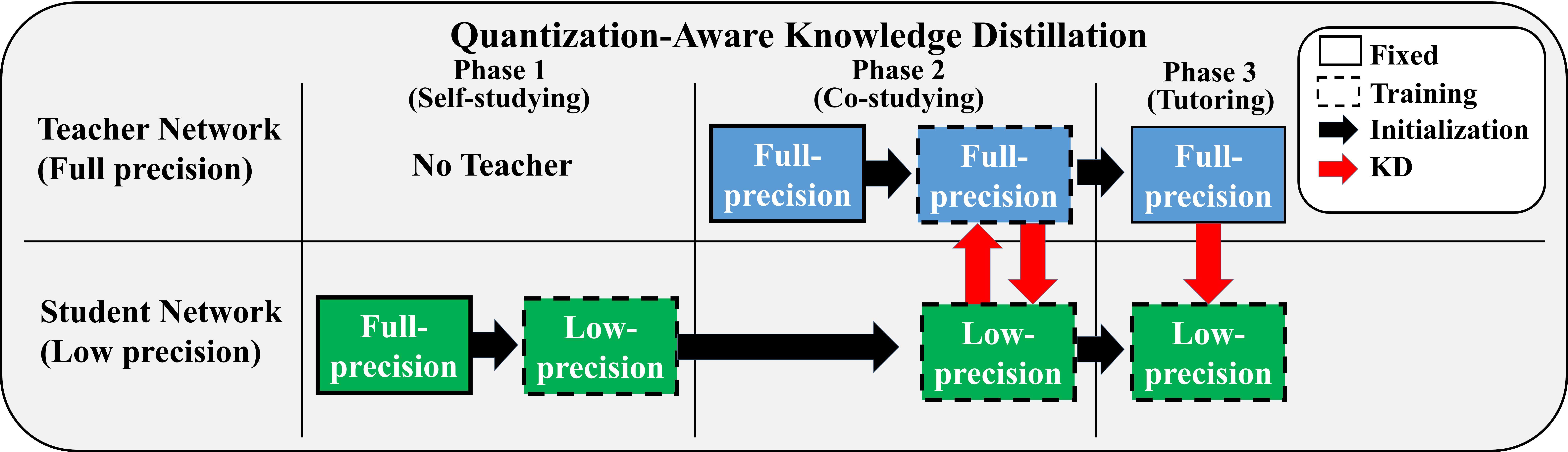}\\
  \caption{The overall process of QKD. Self-studying (SS) phase give a good starting point to alleviate low representative power and regularization effect of KD . Co-studying (CS) phase makes a teacher adaptable to a student and powerful than the fixed teacher. In tutoring (TU) phase, the teacher transfers its adaptable and powerful knowledge to the student.  } % More details are explained in Sec. \ref{method}.}
    \label{fig:overall}
\end{figure*}

In this paper, we propose Quantization-aware Knowledge Distillation (QKD) to address the above mentioned challenges, especially for very low-precision quantization. QKD consists of three phases. {In the first \textbf{`self-studying'} phase, instead of directly applying knowledge distillation to the quantized student network from the beginning, we first try to find a good initialization. In the second \textbf{`co-studying'} phase, we improve the teacher by using the knowledge of the student network. This phase makes the teacher more powerful and quantization-friendly (See Section \ref{discussion}), which is essential to improve accuracy. In final \textbf{`tutoring'} phase, we freeze the teacher network and transfer its knowledge to the student. This phase saves unnecessary training time and memory of the teacher network which tends to have already saturated in the co-studying phase. The overal process of QKD is depicted in the Figure \ref{fig:overall}. Our key contributions are the following: }

%because the teacher tends to be saturated more rapidly than the student in the co-studying step due to its relatively high representative power. 

%We initialize the teacher network with pre-trained full-prescion network and the student network is started with the trained weights in the previous phase. 
% A temperature value is used in the KL divergence term to make both teacher and student distributions softer. Then we transfer their knowledge to each other, which acts as a regularizer for both networks. In the first step of this phase, called co-studying step, we train the teacher and the student jointly with both cross-entropy loss and KL loss. In this step, the teacher is trained to be more adaptable to the student network because the distribution of teacher network is regularized with KL loss by awaring that of the quantized network. 
% Then in the second step called tutoring, we fix the teacher network and 
% transfer its knowledge to the student with only KL loss similar to \cite{mishra2017apprentice}. This saves the training time and memory requirements considerably because the teacher tends to be saturated more rapidly than the student in \nj{the co-studying step} due to its relatively \jh{high representative power}. Our contributions are as the followings:

\begin{itemize}
% \item 
% To effectively combine the contradictory characteristics of low-precision networks' low-representation power and the regularization effect of KD, we propose the phase of `self-studying' as a preprocessing of KD for a better starting point.
%
%\nj{Before applying KD,}
%\jh{For the first phase of QKD, 
%\nj{we} propose `self-studying' \nj{to perform KD} from a better starting point considering \nj{the contradictory} characteristics \nj{of low-precision networks'} low-represatation power  and \nj{the regulization effect} of KD.

% \item In the second phase of QKD, we propose 'Co-studying' and 'Tutoring' (CT), which is suitable for performing quantization and KD, simultaneously.

\item We propose Quantization-aware Knowledge Distillation which can be effectively applied to very low-bit (2-,3-,4-bit) quantized networks. Considering the characteristics of low bit networks and distillation methods, we design a combination of three phases for QKD that overcome the shortcomings posed by conventional quantization + KD methods as described above.

% We propose a new way of combining quantization and KD called QKD which carefully organizes the sequence of quantization and knowledge distillation process. QKD overcomes the accuracy degradation issue of conventional KD+Quantization[][], thereby achieving the state-of-the-art result.

\item We empirically verify that QKD works well on depth-wise convolution networks which are known to be difficult to quantize and our work is the first to apply KD to train low-bit depth-wise convolution networks (viz. MobileNetV2 and EfficientNet)

\item We show that our QKD obtains state-of-the-art accuracies on CIFAR and ImageNet datasets compared to other existing methods. (See Figure \ref{fig:first}.)

%classifier\nj{s} but also \nj{that} of the fused classifier.
 
\end{itemize}

\section{Related work}
% {JW} 

\noindent \textbf{Quantization}\quad Reducing the precision of neural networks \cite{courbariaux2015binaryconnect,li2016ternary,zhu2016trained,nagel2019data,sung2015resiliency} has been studied extensively due to its computational and storage benefits. Although binary $\{-1, +1\}$\cite{courbariaux2015binaryconnect} or tenary $\{-1, 0, +1\}$\cite{li2016ternary} quantizations are typically used, their methods only consider the quantization of weights. To fully utilize bit-wise operations, activation maps also should be quantized in the same way as weights. Some researches consider quantizing both weights and activation maps \cite{hubara2016binarized,soudry2014expectation,rastegari2016xnor,zhou2016dorefa,wang2019haq}. Binary neural networks \cite{hubara2016binarized} quantize both weights and activation maps as binary and compute gradients with binary values. XNOR-Net \cite{rastegari2016xnor} also quantizes its weights and activation maps with scaling factors obtained by constrained optimization. Furthermore, HAQ \cite{wang2019haq} adaptively changes bit-width per layer by leveraging reinforcement learning and HW-awareness.
%predict the bit-width per layer. They quantize each layer with different bit-width .

Recent works have tried to improve quantization further by learning the range of weights and activation maps  %{or the interval values of both %weights and activation maps 
\cite{choi2018pact,jung2019learning,esser2019learned,uhlich2019differentiable}. These approaches can easily outperform previous methods which do not train quantization-related parameters. PACT \cite{choi2018pact} proposes a clipping activation function using trainable a parameter which limits the maximum value of activation. LSQ \cite{esser2019learned} and TQT \cite{tqt2019} introduce uniform quantization using trainable interval values. QIL \cite{jung2019learning} uses a trainable quantizer which performs both pruning and clipping. Our QKD leverages these trainable approaches in the baseline quantization implementation. On top of this quantization-only method, our elaborated knowledge distillation process boosts the accuracy, thereby achieving the state of art accuracy on low-bit quantization.

%We use this simple trainable interval value as our baseline of quantization because uniform quantization is friendly to hardware and
%\nj{it has a better chance of finding good quantization interval values of both weights and activation maps by training the interval values.} %\jh{training the interval values of both weights and activation maps can adaptively \nj{find} the \nj{better} intervals for quantization.} 

\vspace{1mm}

\noindent \textbf{Knowledge Distillation}\quad  KD is one of the most popular methods in model compression. It is widely used in many computer vision tasks. This framework transfers the knowledge of a teacher network to a smaller-sized student network in two ways: Offline and Online. First, offline KD uses a fixed pre-trained teacher networks and the knowledge transfer can happen in different ways. In \cite{hinton2015distilling}, they encourage the student network to mimic the softened distribution of the teacher network. Other works \cite{zagoruyko2016paying,heo2019comprehensive,kim2018paraphrasing,romero2014fitnets} transfer the information using different forms of activation maps from the teacher network. Second, online KD methods \cite{zhu2018knowledge,kim2019feature,zhang2018deep} train both the teacher and the student networks simultaneously without pre-trained teacher models. Deep Mutual Learning (DML) \cite{zhang2018deep} have tried to transfer each knowledge of independent networks using KL loss, although it is not studied in the context of quantization. Our QKD uses both online and offline KD methods sequentially.

\vspace{1mm}
% Due to AP's naive initialization of student network, AP tends to stuck in local mimimum. The self-study of QKD directly mitigate this issue. Furthermore fixed teacher in AP limits the accuracy improvement, which is addressed by co-training in QKD (see Experimental Sectio 5.xx)

\noindent \textbf{Quantization + Knowledge distillation}\quad 
Some researches have tried to use distillation methods to train low precision network \cite{mishra2017apprentice,polino2018model}. They use a full precision network as the teacher and the low bit network as a student. 
%Non-uniform quantization using distillation is proposed in \cite{polino2018model}. 
In Apprentice (AP) \cite{mishra2017apprentice}, the teacher and student networks are initialized with the corresponding pre-trained full precision networks and the student is then fine-tuned using distillation. Due to AP's initialization of the student, AP tends to get stuck in a local minimum in the case of very low-bit quantized student networks. The self-study phase of QKD directly mitigates this issue.
%Initializing with full-precision can not make the full use of KD due to student's low representative power and the regularization effect of KD. To resolve this problem, we propose the self-studying (SS) phase for a good starting point. 
Also, using a fixed teacher, as in  \cite{mishra2017apprentice,polino2018model}, can limit the knowledge transfer due to the inherent differences between the distributions of the full-precision teacher and low-precision student network. We tackle this problem via online co-studying (CS) and offline tutoring (TU). 

% Compared to existing methods of combining quantization and KD, we do not apply KD to the quantized network directly because of the combined detrimental effect arising from the low representation power of low-bit networks and the regularization imposed by KD. To alleviate this issue, we try to find a good starting point by using the so called self-studying phase. Also we do no use a fixed teacher network. Instead, we train the teacher using the knowledge of the student network to make a better teacher compared to a fixed teacher network because in general, teacher-student adaptability %accuracy 
% is very important in transferring meaningful knowledge.

\section{Proposed Method}

In this section, we first explain the quantization method used in QKD and then describe the proposed QKD method.

\subsection{Quantization}
\label{quantizer}
In QKD, we use a trainable uniform quantization scheme for our baseline quantization implementation. We choose uniform quantization because of its hardware-friendly characteristics. In this work, we quantize both weights and activations. So, we introduce two trainable parameters for the interval values of each layer's weight parameters and input activations similar to \cite{esser2019learned,tqt2019,uhlich2019differentiable}. These parameters can be trained with either the task loss or a combination of the task and distillation losses. Considering $k$-bit quantization for weights and activations, the weight and activation quantizers are defined as follows. 

% We use a uniform quantization scheme for our underlying quantization method, where we introduce trainable parameters for the interval value of weights and activations, \nj{similar to}
\vspace{2mm}
\noindent \textbf{Weight Quantizer:} 
Since weights can take positive as well as negative values, we quantize each weight to integers in the range $[-2^{k-1}, 2^{k-1}-1]$. Before rounding, the weights are first constrained to this range using a clamping function $F_W(w) \triangleq F(w, -2^{k-1}, 2^{k-1}-1)$ where
\begin{equation}
F(x, min, max) = \begin{cases}
    max & \text{if } \quad x > max \\
    min & \text{if } \quad x < min \\
    x & \text{elsewise}.
    \end{cases}
    \label{eq:clip}
\end{equation}

%\begin{equation} 
%F_W(w) = hardtanh(w,[-2^{k-1},2^{k-1}-1]).
%\label{eq:clip}
%\end{equation} 

%\cmt{Define hardtanh!!}

We use a trainable parameter $I_W$ for the interval value of weight quantization. It is trained along with the weights of the network. We can calculate the quantization level of the input weight $w$ using a rounding operation on $F_W$. The overall quantization-dequantization scheme for the weights of the network is defined as follows:
\begin{equation}
\hat{w}=Q_W(w) = \Bigl\lfloor{F_W(\frac{w}{I_W})+\frac{1}{2}}\Bigr\rfloor \times I_W 
\label{eq:quan}
\end{equation} 
where $\lfloor \cdot \rfloor$ is the flooring operation. Note that the \textit{dequantization} step (multiplication by $I_W$) just brings the quantized values ($\hat{w}$) back to their original range and is commonly used when emulating the effect of quantization \cite{bhandare2019efficient, tqt2019, rodriguez2018lower}.

\vspace{2mm}
\noindent \textbf{Activation Quantizer:} 
Since most of the networks today use ReLU as the activation function, the activation values are unsigned\footnote{EfficientNet uses Swish extensively which introduces a small proportion of negative activation values, but we can save one bit per activation by clamping these negative values at 0 using unsigned quantization}. Hence, to quantize the activations, we use an quantization function with the range $[0, 2^k-1]$. The activation quantizer $Q_X$ can be obtained as
%\nj{Like ()Using the trainable interval value ($I_X$) of activation and activation quantizer $Q_X$ , we can compute the level ($\hat{x}$) of input activations. 
\begin{equation} 
\hat{x}=Q_X(x) = \Bigl\lfloor{F_X(\frac{x}{I_X})+\frac{1}{2}}\Bigr\rfloor \times I_X
\end{equation} 
where, $F_X(x) \triangleq F(x, 0, 2^k-1)$ and $x,I_X$ represent activation value and the interval value of activation quantization. 

Prior to training, we initialize these interval values ($I_W$,$I_X$) for every layer using the min-max values of weights and activations (form one forward pass), similar to TF-Lite \cite{tensorflow2015-whitepaper}. These quantizers are non-differentiable, so we use a straight-through estimator (STE) \cite{bengio2013estimating} to backprop through the quantization layer. STE approximates the gradient $\frac{d\hat{x}}{dx}$ by 1. Thus, we can approximate the gradient of loss $\mathcal{L}$, $\frac{d\mathcal{L}}{dx}$, with $\frac{d\mathcal{L}}{d\hat{x}}$ 
\begin{equation} 
\frac{d\mathcal{L}}{dx} = \frac{d\mathcal{L}}{d\hat{x}}\frac{d\hat{x}}{dx} \approx \frac{d\mathcal{L}}{d\hat{x}} .
\end{equation}

\subsection{Quantization-aware Knowledge Distillation}
We will now describe the three phases of QKD. Algorithm \ref{algo-training} depicts the overall QKD training process.

%\yb{Here, we describe our proposed} quantization-aware knowledge distillation (QKD) method which can be effectively applied to very low bit (2,3 and 4 bit) quantization. 
%We consider the characteristics of low bit networks and distillation methods together \yb{to formulate three phases for QKD}. In terms of KD, we use both online (training teacher) and offline (freezing teacher) distillation methods for QKD. \yb{Algorithm \ref{algo-training} depicts the overall QKD training process which we describe below.}

% On the first phase (Self-studying), Considering the low representative power of low bit network and regularizing effect of KD, we train only low-bit network with task loss for finding good starting point. At the second phase (Co-studying), we try to use extra knowledge from powerful teacher compared to fixed teacher so jointly train the teacher and student to make powerful teacher than fixed teacher for transferring strong knowledge to the student. In the third phase (Tutoring), we use the rapidly saturated teacher and fix it to reduce computational cost and memory for calculating the gradient teacher. Then, we use this fixed strong teacher to train the student. 

\subsubsection{Phase 1: Self-studying}

Directly combining quantization and KD cannot make the most of the positive effects of KD and easily cause unexpectedly lower performance. This is because of the regularizing characteristics of KD and limited representative power of the quantized network. This can cause the quantized network to easily get trapped in a poor local minima. To mitigate this issue and to provide a good starting point for KD, we train the low-bit network for some epochs using only the task loss, i.e. the standard cross entropy loss. Such a self-studying phase provides the student with good initial parameters before KD is applied.

Our method can be compared to progressive quantization (PQ) \cite{zhuang2018towards} which also uses two stages and progressively quantizes weights and activation maps for a good initial point. While PQ conducts progressive quantization which uses a higher precision network as a initialization and conducts iterative update between weight and activation maps, our method directly initializes the same low bit-width for the target low-bit network because learning the interval of weights and activation maps works well without iterative and progressive training.

This strategy of parameter initialization and knowledge distillation has good synergy in terms of generalization and finding a good local minimum. More specifically, our initialization scheme helps to start with a good starting point and the distillation loss guides the student network to good local minima acting as a regularizer.

% Our method can be categorized as a fine-tuning method similar to the third scheme of \yb{Apprentice} \cite{mishra2017apprentice} which initializes the teacher and the student network with a pre-trained full precision network. However, differently from \cite{mishra2017apprentice}, in our case, before distilling the knowledge of the teacher network, we train the low-bit network using only the task loss such as cross entropy. Such a self-studying phase provides the student with good initial parameters before KD is applied. Then, we use a pretrained full precision network as the teacher and the low bit network from the self-studying phase as the student in the knowledge distillation phase. % \jh{which contains co-studying and tutoring}%After initialization, we use distillation loss for transferring the knowledge of the teacher network. 

\begin{algorithm}[t]
%%\label{alg:1}
\renewcommand{\algorithmicrequire}{\textbf{Input:}}
\renewcommand{\algorithmicensure}{\textbf{Output:}}
\renewcommand{\algorithmicprint}{\textbf{break}}
\caption{\text{ Quantization-aware Knowledge Distillation}}
\label{algo-training}
\begin{algorithmic}[1]
\REQUIRE Training data; \\ Pre-trained FP weights for teacher model $T_F$; \\ Pre-trained FP weights for student model $S_F$; \\ Low-bit student model weights ${S_L}$; \\ Weight interval values $I_W$; \\ Activation interval values $I_X$; \\ Number of epochs for each phase $P_1$; $P_2$; $P_3$;
\ENSURE Trained low-bit student weight and interval values ${S_L}^\prime$, ${I_W}^\prime$ and ${I_X}^\prime$ %which are trained with teacher model $T_F$.
% Phase 1
\STATE \textbf{Phase 1: Self-studying} 
\STATE Init $S_L$ with $S_F$; Init $I_W,I_X$ using min-max values of weights and one batch of activations;
\FOR{ Epoch = 1 ,..., $P_1$}
\STATE Update $S_L$, $I_W$ and $I_X$ by minimizing $\mathcal{L}^{s}_{ce}$ (\ref{eq:Lce}) %using backpropagation
\ENDFOR

\STATE \textbf{Phase 2: Co-studying} 
\STATE Init ${S_L}^\prime$, ${I_W}^\prime$ and ${I_X}^\prime$ with $S_L$, $I_W$ and $I_X$%; 
%Initialize ${I_W}^\prime$ and ${I_X}^\prime$ \nj{with} $I_W$ and $I_X$
\FOR{ Epoch = 1 ,..., $P_2$} 

\STATE Update $T_F$ by minimizing $\mathcal{L}^{T}_{KD}$ (\ref{eq:LKDT})%using backpropagation

\STATE Update ${S_L}^\prime$, ${I_W}^\prime$ and ${I_X}^\prime$ by minimizing $\mathcal{L}^{S}_{KD}$ (\ref{eq:LKDS})%using backpropagation

\ENDFOR

\STATE \textbf{Phase 3: Tutoring}
\FOR{ Epoch = $1$ ,..., $P_3$} 
\STATE Update ${S_L}^\prime$, ${I_W}^\prime$ and ${I_X}^\prime$ by minimizing $\mathcal{L}^{S}_{KD}$ (\ref{eq:LKDS})%using backpropagation
\ENDFOR

\end{algorithmic}
\end{algorithm}

\subsubsection{Phase 2: Co-studying}

% Offline and online KD methods have their pros and cons. Unlike online methods, in offline methods, %\cmt{both online and offline methods have this feature?}, 
% one can use not only the output class probabilities but also meaningful feature maps in the training of the student. Even though offline methods can use feature maps, they need pre-trained fixed teachers %and \nj{they use fixed teacher so it 
% that they can only utilize information limited to fixed teachers. On the other hand, online methods train both the student and the teacher simultaneously. %not only train the student network but also train the teacher network. 
% In doing so, they can utilize better information than that of fixed teacher networks because teachers can also be trained with knowledge of student. However, these methods need to calculate the gradient of teacher networks requiring more memory and training time than offline methods. 

% In the general knowledge distillation domain, a powerful teacher with high accuracy can teach student better than a weaker one \cite{zhu2018knowledge}. \jh{Also, It is important to make teacher adaptable to the student because the inherent gap between the teacher and student hinder to train student \cite{kim2018paraphrasing,mirzadeh2019improved} Previous works \cite{mishra2017apprentice,polino2018model} use a fixed teacher which has limitations in terms of powerful and adaptable information that can be transferred for training a low-bit student network. They do not consider above two characteristic of KD.}

To make a powerful and adaptable teacher, we jointly train the teacher network (full-precision) and the student network (low-precision) in an online manner. Kullback–Leibler divergence (KL) between the student and teacher distributions is used to the make the teacher more powerful in terms of accuracy as well as it's adaptability to the student distribution than the fixed pre-trained teacher. In this framework, teacher network is trained by softened distribution of student network and vice versa. Teacher network can be adapted to the quantized student network with KL loss by awaring the distribution of the quantized network. 

Assuming that there are $m$ classes, the cross-entropy loss for \textit{both} the teacher and the student networks is obtained by firstly computing the softmax posterior with temperature $\mathcal{T}$ as follows:
%We can calculate the \nj{cross-entropy using the posterior for both the teacher and the student networks} :
\begin{equation} 
p_i(\mathbf{z}^k;\mathcal{T}) = \frac{ e^{z^k_i/\mathcal{T}}}{\sum_{j}^me^{z^k_j/\mathcal{T}}},
\end{equation} 
\begin{equation} 
\mathcal{L}_{ce}^k = -\sum_{i=1}^{m}y^{(i)}\log(p_i(\mathbf{z}^k;1)),
\label{eq:Lce}
\end{equation} 
where
$\mathbf{z}^k$ and $\mathcal{L}_{ce}^k$ represent logit and cross-entropy of the $k$-th network, i.e. the student or the teacher ($k = \{S,T\}$). The temperature value, $\mathcal{T}$, is used to make distribution softer for using the dark knowledge. We can compute the KL loss between student and teacher network using logits. 
\begin{equation} 
{KL}(z^T||z^S; \mathcal{T}) = \sum_{i=1}^{m}p_i(z^T;\mathcal{T})\log(\frac{p_i(z^T;\mathcal{T})}{p_i(z^S;\mathcal{T})}).
\label{eq:KLloss}
\end{equation} 

 Then, we update each network with cross entropy and KL loss as below:
\begin{equation} 
\mathcal{L}^{S}_{KD} = \mathcal{L}_{ce}^S + \mathcal{T}^2\times{KL}(z^T||z^S; \mathcal{T})
\label{eq:LKDS}
\end{equation} 
\begin{equation} 
\mathcal{L}^{T}_{KD} = \mathcal{L}_{ce}^T + \mathcal{T}^2\times{KL}(z^S||z^T; \mathcal{T})
\label{eq:LKDT}
\end{equation} 
$\mathcal{L}^{S}_{KD}$ and $\mathcal{L}^{S}_{KD}$ refer to the loss of student and teacher network, respectively. We multiply $\mathcal{T}^2$ because the gradient scale of logits decrease as much as $1/\mathcal{T}^2$.

\subsubsection{Phase 3: Tutoring}

After a few epochs of co-studying, the accuracy of teacher network starts saturating. This is because the teacher (full-precision) has relatively high representative power than the student (low-precision). To reduce the computational cost and memory in calculating the gradient of the teacher network, we freeze the teacher network and train only the low-bit student network with $\mathcal{L}^{S}_{KD}$ loss in an offline manner. In this phase, we use the knowledge of a teacher network which is now more quantization-aware as a result of co-studying. As we will show later (See Section \ref{discussion}), using tutoring gives us equal or better student performance compared to only using co-studying throughout the training.

% Moving this line up...
% The overall training process of QKD is depicted in Algorithm \ref{algo-training}.   

\section{Experiments}
We perform a comprehensive evaluation of our method on the CIFAR10/100 \cite{krizhevsky2009learning} and ImageNet \cite{ILSVRC15} datasets.
We compare the proposed QKD with existing state-of-the-art quantization methods on 2, 3, and 4-bits (i.e., W2A2, W3A3, W4A4). To show the robustness of our method, we provide comparisons on standard convolutions (i.e., ResNet \cite{he2016deep}) as well as depth-wise separable convolutions (i.e., MobileNetV2 \cite{sandler2018mobilenetv2} and EfficientNet \cite{tan2019efficientnet}). Furthermore, we perform an extensive ablation study to analyze the effectiveness of the different components of QKD. Following methods are considered for performance comparison:

%We compare the proposed QKD with existing state-of-the-art quantization methods on 2-bit (W2A2), 3-bit (W3A3) and 4-bit (W4A4) quantization. To show the effectiveness of our method on quantizing different types of convolutions, we provide comparisons on the original ResNet architectures \cite{he2016deep} with standard convolutions as well as on MobileNetV2 \cite{sandler2018mobilenetv2} and EfficientNet \cite{tan2019efficientnet} with depth-wise separable convolutions. 
%Furthermore, we provide an ablation study of the contributions mentioned in this work. 

% Quantizing architectures with depth-wise separable convolutions is challenging. It has been observed that quantizing MobileNet even to W6A6 or lower bit-widths leads to severe degradation in performance \cite{krishnamoorthi2018quantizing, nagel2019data}. One of the reasons is that these architectures are optimized for inference, hence there is little room for further compression or quantization. We observed severe degradation in accuracy on applying these quantization methods to EfficientNet and MobileNetV2. In section \ref{depthwise}, we show that QKD offers a solution to alleviate this loss in accuracy to a certain extent.

%In the experimental results and discussion (Section \ref{discussion}), we perform an extensive ablation study to analyze the effectiveness of the different components of QKD. For comparison with the existing methods, we consider the following:
\begin{enumerate}[leftmargin=*]
    \itemsep0pt
    \item `\textbf{Baseline (BL)}' is the baseline quantization-only version of QKD as described in \ref{quantizer}; no teacher is used during training. We train the low precision network using cross-entropy and initialize the weights with the pretrained full-precision weights.
    %\item `\textbf{Baseline (BL)}' is the underlying quantization method for QKD that was described in Sec \ref{quantizer}. This is a quantization-only method, meaning that no teacher is used during training. We train the low precision network using cross-entropy and initialize the weights with the pretrained full-precision weights.
    %In training BL, there is no teacher and we train BL only using cross-entropy by initializing the weight of BL with pre-trained full-precision network.

    \item `\textbf{SS + BL}' is BL, but the low-bit betwork is initialized with the weights and interval values trained in the self-studying (SS) phase.\footnote{Note that SS is the same as BL, only difference being that during SS, the low-bit network is trained for much fewer epochs.}

    \item `\textbf{AP*}' is a modified version of the original Apprentice \cite{mishra2017apprentice} method for knowledge distillation. The original Apprentice uses WRPN scheme \cite{mishra2017wrpn}. We replace this quantizer with BL and initialize both teacher and student with pre-trained full-precision newtork. %\vspace{2mm}\\ 

    \item `\textbf{SS + AP*}' is AP* initialized with the weights and interval values trained in the self-studying (SS) phase for the low-precision network. %\vspace{2mm}\\ 

    \item `\textbf{CS + TU}' means that we initialize the student network with pre-trained full-precision network the same way as AP*, then perform ``Co-studying" and ``Tutoring"  between the teacher and the student using BL. 

    \item `\textbf{QKD (SS + CS + TU)}' is our proposed method. We initialize the student network by the SS phase. Then, we perform KD in the CS + TU phase.
\end{enumerate}

\subsection{Implementation Details}
We quantize the weights and input activations of all the Conv and Linear layers as described in Section \ref{quantizer}. We quantize first and last layer to 8-bits to ensure compatibility on any fixed-point hardware. We set the temperature value $\mathcal{T}$ as 2 in QKD. In all the experiments, we use the same settings while training all 6 baselines mentioned above. 
% No need to say the following sentence... Sounds a little arrogant...
%Comparing other method, they do not quantize first and the last layer because of accuracy degradation.

\vspace{1mm}

\noindent \textbf{CIFAR}\quad We train for up to $200$ epochs with step learning rate schedule with SGD same as \cite{zhang2018lq}. We use the starting learning rate (LR) of $0.1$ and $0.01$ for models corresponding to CIFAR-10/100 respectively. We use 30 epochs for SS phase by dividing LR with $10$ for every 10 epoch. After SS phase, we reset the LR to $0.1$ and $0.01$ corresponding to CIFAR-10/100. Then, we use other 170 epochs for CS + TU phase. LR learning rate is divided by $10$ at $80$ and $120$ epoch. We use 100 epochs for CS and the remaining 70 epochs for TU. For teacher network, we use the same schedule as the student for the CS phase, but we freeze the teacher in TU. Compared to the LR used for student model's weight parameters, we use 100 times smaller LR for updating $I_W$ and same LR for updating $I_X$.

\vspace{1mm}

\noindent \textbf{ImageNet}\quad For all our compared methods, we used total 120 epochs for training (this is the same length as QIL \cite{jung2019learning}). In the methods SS + BL, SS + AP* and SS + CS + TU, we finetune the student model for 50 epochs in the SS phase and the remaining 70 epochs are used for the rest. In the CS + TU method, the first 60 epochs are for CS and the remaining 60 epochs for TU. For training the student, we used a Mixed Optimizer (MO) setting \cite{opennmt}. In the MO setting, we use SGD \cite{bottou2010large} optimizer to update the weights and Adam \cite{kingma2014adam} for updating the interval values ($I_W$, $I_X$). We found that the exponential learning rate schedule works best for QKD. For the ResNet architectures, we use an initial learning rate of $0.001$ for both SS and CS + TU phases. For EfficientNet and MobileNetV2, initial learning rate of $0.003$ was used. Similar to CIFAR-100 setting, we use 100 times smaller LR for $I_W$ and same LR for $I_X$, compared to LR for model weights. For $I_W$ and $I_X$, not much fine-tuning was required because of the Mixed Optimizer setting. More details are described in supplementary details.

\subsection{CIFAR-10 Results}
We evaluate our method on the CIFAR-10 dataset. The performance numbers are shown in the Table \ref{table:cifar10}. We use ResNet-56 (FP : 93.4\%) as a teacher network for both AP* and QKD. Interestingly, for AP* and CS + TU, which combine BL and knowledge distillation, performance is worse than BL at W2A2. This is because the network has low-representative power at W2A2 and can be negatively affected by the heavy regularization imposed by KD. To alleviate this issue, we use the weights trained in the SS phase for a better initialization for the weights of the student. SS + AP* and QKD outperform the existing methods and the Baseline (BL) method for all the cases. These results suggest that initialization is very important in combining quantization and \textit{any} KD method. In our proposed method, 2\% gain was observed at W2A2 with the SS phase. Compared to the distillation method AP*, QKD which also trains the teacher shows a significant improvement in performance. QKD also provides best results for W3A3 and W4A4.
\begin{table}[t]
	\caption{CIFAR-10 top-1 accuracy}
% 	\small
	\centering
	\begin{adjustbox}{width=0.8\linewidth}
	\begin{tabular}{p{1.6cm}| c| c c c}
        \toprule
        	{Network} & {Method} & {W2A2} & {W3A3} & {W4A4} \\
			
			\midrule
	
			\multirow{8}{1.6cm}{ResNet-20 \\ (FP : 91.6)} & DoReFa-Net \cite{zhou2016dorefa} & 88.2 & 89.9 & 90.5 \\
			& PACT \cite{choi2018pact} & 89.7 & 91.1 & 91.3  \\
			& LQ-Net \cite{zhang2018lq} & 90.2 & 91.6 & -- \\ 
			\cline{2-5}
			& BL & 89.9 & 91.8 & 92.1 \\
			& SS + BL  & 90.1 & 91.9 & 92.3  \\
			& AP* & 88.6 & 91.8 & 92.2  \\
			& SS + AP*  & 90.2 & 92.3 & 92.3 \\
			& CS + TU & 88.5 & 92.1 & 92.5 \\
			& QKD & \textbf{90.5} & \textbf{92.7} & \textbf{93.1}  \\
        \bottomrule
	\end{tabular}
	\end{adjustbox}
	\label{table:cifar10}
\end{table}

\subsection{CIFAR-100 Results}
Table \ref{table:cifar100} shows the experimental results on CIFAR-100. We use a ResNet-56 (FP : 73.4\%) as a teacher in this experiment. These experiments show similar tendencies to those with CIFAR-10. The accuracy increases with the SS phase which helps us start with a better initialization especially in very low-bit quantization (W2A2). Considering KD methods, QKD performs significantly better than SS+AP*. We attribute this gain in accuracy to the use of an adaptable teacher trained during co-studying compared to a fixed teacher. Further discussion on this is provided in Section \ref{discussion}. Compared to the full-precision version of ResNet-32, the W3A3 and W4A4 quantized versions have a 1.4\% and 2.5\% gain respectively.

\begin{table}[t]
	\caption{CIFAR-100 top-1 accuracy}
% 	\small
	\centering
	\begin{adjustbox}{width=0.8\linewidth}
	\begin{tabular}{p{1.6cm}| c| c c c}
        \toprule
        	{Network} & {Method} & {W2A2} & {W3A3} & {W4A4} \\
			
			\midrule
	
			\multirow{5}{1.6cm}{ResNet-32 \\ (FP : 70.8)} & BL & 65.4 & 69.8 & 70.9 \\
			& SS + BL  & 65.7 & 70.5 & 70.8  \\
			& AP* & 63.5 & 70.3 & 71.5  \\
			& SS + AP*  & 66.1 & 70.6 & 71.6 \\
			& CS + TU & 62.3 & 71.8 & 73.2 \\
			& QKD  & \textbf{66.4} & \textbf{72.2} & \textbf{73.3}  \\
        \bottomrule
	\end{tabular}
	\end{adjustbox}
	\label{table:cifar100}
\end{table}

\subsection{ImageNet Results}
%For ImageNet, we show the performance of our method on architectures with both standard and depthwise separable convolutions. For standard convolutions, Table \ref{table:standard} shows the results on the ResNet family of architectures. For depthwise convolutions, Table \ref{table:depthwise} and show results on MobileNetV2 and EfficientNet-B0. 

\subsubsection{Standard Convolutions}
Table \ref{table:standard} shows the performance of the proposed method on the original ResNet18, ResNet34 and ResNet50 architectures \cite{he2016deep}. We compare our proposed QKD method with QIL \cite{jung2019learning}, PACT \cite{choi2018pact}, DSQ \cite{gong2019differentiable} and LQ-Nets \cite{zhang2018lq}, which are current SOTA methods that show results with 2-, 3-, and 4-bit quantization on the original ResNet architectures. We use ResNet101, ResNet50 and ResNet34 as teachers for the student networks ResNet50, ResNet34 and ResNet18 respectively.

We observed that the QKD method outperforms the existing approaches in terms of both top-1 and top-5 accuracy. Our distillation method consistently gave us 0.5-1.1\% gain in top-1 accuracy across all bit-widths over our baseline quantization method (BL). QKD even exceeds the full-precision accuracy by more than $1\%$ at W4A4 for all ResNet architectures.
Interestingly, CS+TU outperforms AP* everywhere but SS+AP* has better performance than CS+TU at W2A2. This again confirms the efficacy of using self-studying especially at 2-bit quantization. Also, it can be seen that QKD (SS+CS+TU) outperforms all the other methods at W2A2.

% We note that QKD method can even match the full-precision performance with W3A3 quantization and get a 0.4\% gain with ResNet34. \jh{Even at W4A4, the accuracy of QKD is bout 1\% more better than that of full-precision (W32A32).} Note that, KD also has a regularization effect during training and this is noticeable in the case of W2A2 quantization - it can be seen that the CT (co-studying and tutoring) performs worse than SS+AP (Apprentice with Self-Studying based initialization). But with the SS initialization, QKD is able to get 0.3\% gain over SS+AP in the W2A2 case.

\begin{table}[t]
	\caption{ImageNet performance comparison on architectures with standard convolutions}
% 	\small
	\begin{adjustbox}{width=1\linewidth}
	\begin{tabular}{c| c| c c |c c | c c}
        \toprule
        	\multirow{2}{*}{Bitwidth} & \multirow{2}{*}{Method} & \multicolumn{2}{c}{ResNet18} & \multicolumn{2}{c}{ResNet34} &
        	\multicolumn{2}{c}{ResNet50} \\
			
			\cmidrule(lr){3-4}
			\cmidrule(lr){5-6}
			\cmidrule(lr){7-8}

			 & & top-1 & top-5 & top-1 & top-5 & top-1 & top-5 \\
            
			\addlinespace[0.2em]
            \midrule
            
            W32A32 & - & 70.1 & 89.5 & 73.5 & 91.7 & 76.3 & 93.0 \\
            
            \addlinespace[0.2em]
            \midrule
			\multirow{8}{*}{W4A4} 
			%&  DoReFa-Net \cite{zhou2016dorefa} & 68.1 & -- & -- & -- & 71.4 & -- \\
			
			& PACT \cite{choi2018pact} & 69.2 & 89.0 & - & - & 76.5 & 93.2 \\
			& LQ-Nets \cite{zhang2018lq} & 69.3 & 88.8 & - & - & 75.1 & 92.4 \\
			& DSQ \cite{gong2019differentiable} & 69.6 & - & 72.8 & - & - & - \\
			& QIL \cite{jung2019learning} & 70.1 & - & 73.7 & - & - & -\\ \cline{2-8}

			& BL & 70.7 & 89.4 & 73.8 & 91.5 & 76.6 & 93.2 \\
			& SS + BL & 70.8 & 89.4 & 74.9 & 91.5 & 76.6 & 93.2 \\
			& AP* & 70.9 & 89.5 & 74.0 & 91.7 & 76.7 & 93.2 \\
			& SS + AP* & 71.1 & 89.6 & 74.2 & 91.8 & 76.9 & 93.3 \\
			& CS + TU & 71.1 & 90.1 & 74.4 & 91.9 & 77.0 & 93.5 \\
			& QKD  & \textbf{71.4} & \textbf{90.3} & \textbf{74.6} & \textbf{92.1} & \textbf{77.3} & \textbf{93.6} \\
            
            \midrule
			\multirow{8}{*}{W3A3} 
			%&  DoReFa-Net \cite{zhou2016dorefa} & 67.5 & -- & -- & -- & 69.9 & -- \\

			& PACT \cite{choi2018pact}& 68.1 & 88.2 & - & - & 75.3 & 92.6 \\
			& LQ-Nets \cite{zhang2018lq}& 68.2 & 87.9 & 71.9 & 90.2 & 74.2 & 91.6 \\
			& DSQ \cite{gong2019differentiable} & 68.7 & - & 72.5 & - & - & - \\
			& QIL \cite{jung2019learning} & 69.2 & - & 73.1 & - & - & - \\ 
			\cline{2-8}
			
			& BL & 69.4 & 88.5 & 73.3 & 90.5 & 75.3 & 92.6 \\
			& SS + BL & 69.5 & 88.5 & 73.3 & 90.6 & 75.4 & 92.7 \\
			& AP* & 69.7 & 88.7 & 73.5 & 90.7 & 76.0 & 92.7 \\
			& SS + AP* & 69.9 & 88.7 & 73.7 & 90.8 & 76.1 & 92.8 \\
			& CS + TU & 70.0 & 89.4 & 73.7  & 91.2 & 76.2 & 92.9 \\
			& QKD  & \textbf{70.2} & \textbf{89.9} & \textbf{73.9}  & \textbf{91.4} & \textbf{76.4} & \textbf{93.2} \\
			\midrule
			\addlinespace[0.2em]

			\multirow{8}{*}{W2A2} 
			%&  DoReFa-Net \cite{zhou2016dorefa} & 62.6 & -- & -- & -- & 67.1 & -- \\

			& PACT \cite{choi2018pact} & 64.4 & 85.6 & - & - & 72.2 & 90.5 \\
			& LQ-Nets \cite{zhang2018lq}& 64.9 & 85.9 & 69.8 & 89.1 & 71.5 & 90.3 \\
			& DSQ \cite{gong2019differentiable}& 65.2 & - & 70.0 & - & - & - \\
			& QIL \cite{jung2019learning}& 65.7 & - & 70.6 & - & - & -\\ \cline{2-8}
			
			& BL & 66.7 & 86.7 & 71.0 & 89.8 & 73.3 & 91.0 \\
			& SS + BL & 66.8 & 86.9 & 71.0 & 89.9 & 73.3 & 91.0 \\
			& AP* & 66.8 & 86.9 & 71.1 & 89.9 & 73.4 & 91.0 \\
			& SS + AP* & 67.2 & 87.3 & 71.3 & 90.2 & 73.7 & 91.2 \\
			& CS + TU & 67.1 & 87.2 & 71.3 & 90.1 & 73.6 & 91.2 \\
			& QKD  & \textbf{67.4} & \textbf{87.5} & \textbf{71.6} & \textbf{90.3} & \textbf{73.9} & \textbf{91.6} \\
        \bottomrule
	\end{tabular}
	\end{adjustbox}
	\label{table:standard}
\end{table}

\begin{table}[t]
	\caption{ImageNet performance comparison on architectures with depthwise separable convolutions}
% 	\small
	\centering
	\begin{adjustbox}{width=0.8\linewidth}
	\begin{tabular}{c | c | c c | c c}
        \toprule
        	\multirow{2}{*}{Bitwidth} & \multirow{2}{*}{Method} & \multicolumn{2}{c}{MobileNetV2} & \multicolumn{2}{c}{EfficientNet-B0} \\
			
			\cmidrule(lr){3-4}
			\cmidrule(lr){5-6}

			 & & top-1 & top-5 & top-1 & top-5 \\
			 
 			\addlinespace[0.2em]
            \midrule
            
            W32A32 & - & 71.8 & 90.3 & 76.3 & 93.2 \\

            \midrule
			\addlinespace[0.2em]
			\multirow{7}{1cm}{W6A6} & PACT \cite{choi2018pact} & 71.3 & 90.0 & - & - \\ \cline{2-6}
			& BL & 71.2 & 89.9 & 74.5 & 92.3 \\
			& SS + BL & 71.3 & 90.0 & 74.6 & 92.3 \\
			& AP* & 71.4 & 90.0 & 74.8 & 92.4 \\
			& SS + AP* & 71.4 & 90.0 & 75.0 & 92.4 \\
			& CS + TU & 71.5 & 90.1 & 75.1 & 92.6 \\
			& QKD  & \textbf{71.8} & \textbf{90.2} & \textbf{75.4} &  \textbf{92.6} \\

            \midrule
			\addlinespace[0.2em]
			\multirow{8}{1cm}{W4A4} & PACT \cite{choi2018pact} & 61.4 & 83.7 & - & - \\ 
			& DSQ \cite{gong2019differentiable} & 64.8 & - & - & -\\ \cline{2-6}
			& BL & 66.1 & 86.2 & 71.9 & 90.4 \\
			& SS + BL & 66.3 & 86.3 & 72.0 & 90.4 \\
			& AP* & 66.4 & 86.4 & 72.3 & 90.6 \\
			& SS + AP* & 66.6 & 86.5 & 72.6 & 90.7 \\
			& CS + TU & 66.7 & 86.7 & 72.8 & 91.0 \\
			& QKD  & \textbf{67.4} & \textbf{87.0} & \textbf{73.1} & \textbf{91.2} \\
			
			\midrule
			\addlinespace[0.2em]
            
			\multirow{6}{1cm}{W3A3} & BL & 60.1 & 83.0 & 67.5 & 85.8 \\
			& SS + BL & 61.5 & 83.4 & 67.7 & 85.9 \\
			& AP* & 60.7 & 83.2 & 68.4 & 86.1 \\
			& SS + AP* & 62.0 & 83.5 & 68.7 & 86.2 \\
			& CS + TU & 62.0 & 83.5 & 68.8 & 86.4 \\
			& QKD  & \textbf{62.6} & \textbf{84.0} & \textbf{69.2} & \textbf{86.9} \\

			\midrule
			\multirow{6}{1cm}{W2A2} 
			& BL & 37.7 & 64.1 & 43.5 & 67.5 \\
			& SS+BL & 42.7 & 66.2 & 44.9 & 67.6 \\
			& AP* & 39.8 & 65.3 & 46.4 & 68.2 \\
			& SS+AP* & 43.9 & 67.3 & 47.7 & 68.5 \\
			& CS + TU & 43.6 & 66.9 & 48.0 & 68.7 \\
			& QKD  & \textbf{45.7} & \textbf{68.1} & \textbf{50.0} & \textbf{69.3} \\

        \bottomrule
	\end{tabular}
	\end{adjustbox}
	\label{table:depthwise}
\end{table}

\subsubsection{Depthwise Separable Convolutions}
\label{depthwise}
%To see the effectiveness of our approach on depth-wise convolutions, we evaluate our method on 
Table \ref{table:depthwise} shows the performance of our method on MobileNetv2 (width\_multiplier=$1.0$) \cite{sandler2018mobilenetv2} and EfficientNet-B0 \cite{tan2019efficientnet}. DSQ \cite{gong2019differentiable} shows results on MobileNetV2 at W4A4. We refer to the PACT \cite{choi2018pact} performance on MobileNetV2 from the HAQ paper \cite{wang2019haq}.  EfficientNet-B0 is the current state-of-the-art on ImageNet among architectures with similar parameter and MAC count. Hence, we also provide the results of our method on EfficientNet-B0. We include W6A6 into our set of bit-widths. We use MobileNetV2 (width=$1.4$) as teacher for the student network MobileNetV2 (width=$1.0$) and EfficientNet-B1 as the teacher for EfficientNet-B0.

In general, we see higher gains with the QKD (over the BL and AP* methods) with both MobileNetV2 and EfficientNet-B0 than what we observed on the ResNet architectures. If we compare a similar accuracy range (66$\sim$67\%) which is observed with W4A4 on MobileNetV2 and W3A3 on EfficientNet-B0, we can see 1.2\% and 1.7\% top-1 accuracy improvement respectively with QKD over BL. Interestingly, we observe a significant difference of 1.3\% between AP* and SS+AP* at W3A3 with MobileNetV2.

With W2A2, we observe a drastic drop in top-1 accuracy. This was expected since depth-wise convolution layers are known to be highly sensitive to quantization \cite{wang2019haq, tqt2019}. So, we ran another set of experiments where the weights and input activations of the depth-wise separable layers are quantized to W8A8 whereas the rest of the layers are quantized to W2A2, W3A3 or W4A4. We include these results in the supplementary details.

 \begin{figure}[t]
 	\centering
 	\includegraphics[width=0.6\linewidth]{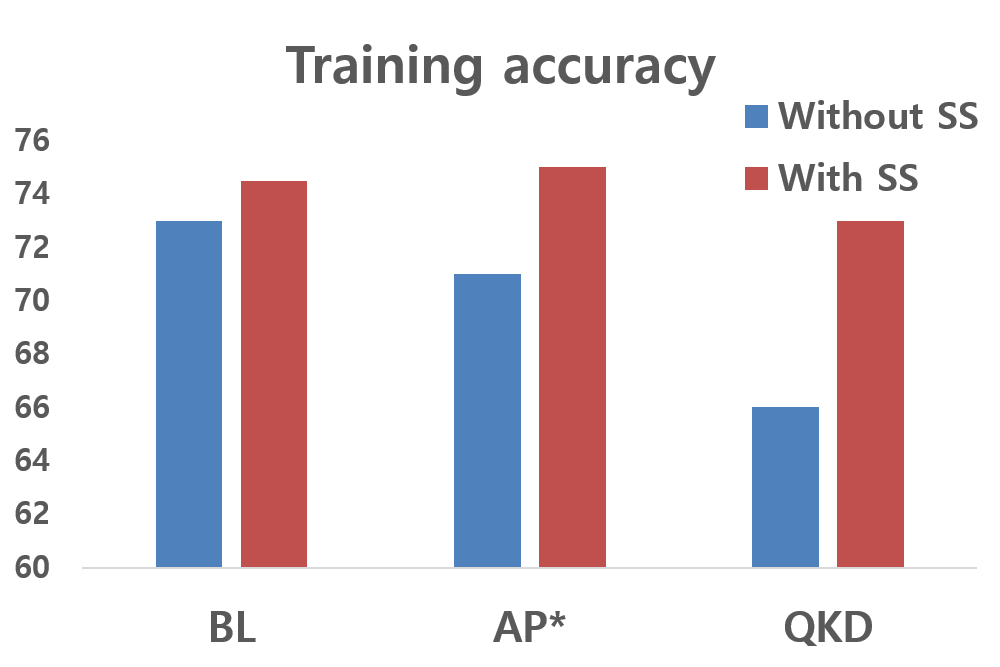}
 	\caption{Training accuracy of student network using BL, AP* and QKD at W2A2 on CIFAR-100. Red and blue represent the accuracy with and without Self-studying (SS)}
 	\label{fig:Tacc}
 \end{figure}

\vspace{1mm}

 \begin{figure}[h]
 	\centering
 	\includegraphics[width=0.8\linewidth]{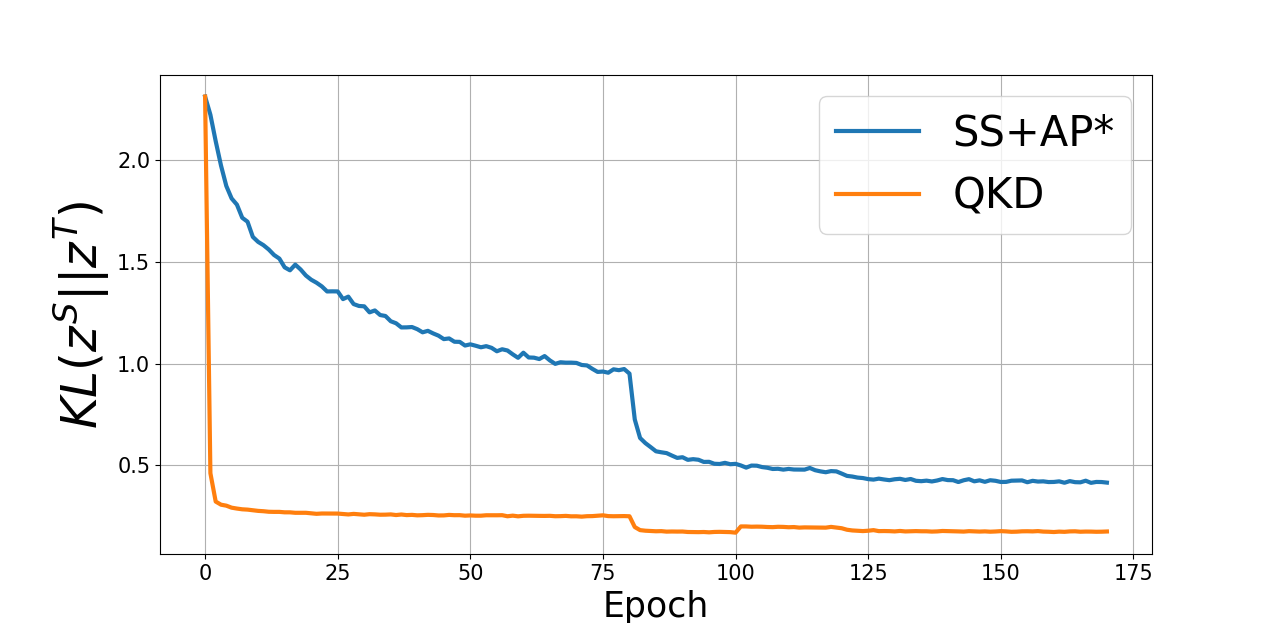}
 	\caption{\small{KL-divergence between teacher \& student distributions on CIFAR100 with QKD and SS+AP* (SS part is not shown in both because it doesn't involve the teacher network).}}
 	\label{fig:KL}
 \end{figure}

\vspace{1mm}

\begin{figure*}[t]
  \begin{subfigure}[b]{0.33\textwidth}
    \includegraphics[width=\textwidth]{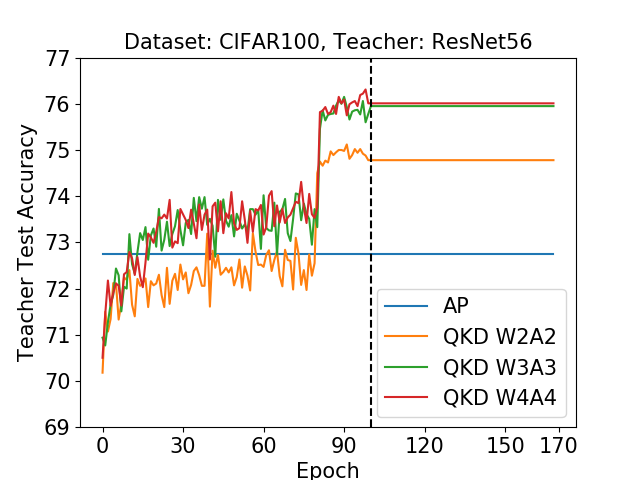}
  \end{subfigure}
  \begin{subfigure}[b]{0.33\textwidth}
    \includegraphics[width=\textwidth]{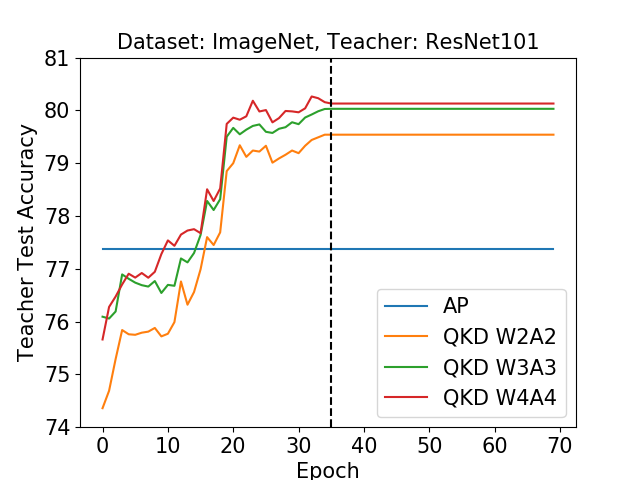}
  \end{subfigure}
  \begin{subfigure}[b]{0.33\textwidth}
    \includegraphics[width=\textwidth]{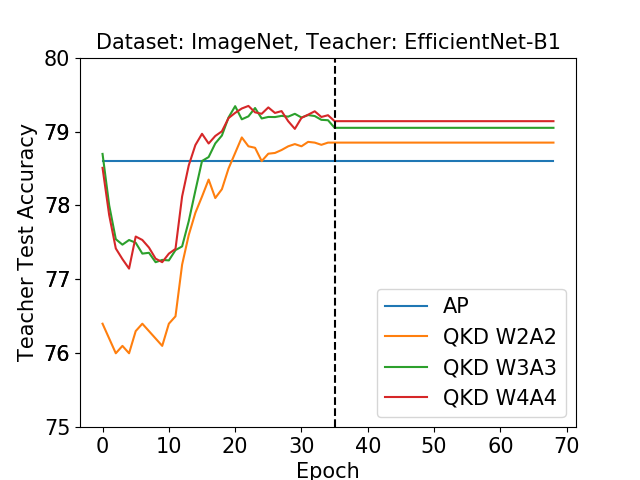}
  \end{subfigure}
  \caption{Accuracy of the teacher network during the CS + TU phase for 3 different scenarios. The vertical dotted line separates Co-studying and Tutoring phases. The blue horizontal line shows the teacher accuracy for Apprentice (AP*) method.}
  \label{fig:teacheracc}
\end{figure*}

\subsection{Discussion }
\label{discussion}

\noindent \textbf{Effectiveness of Self-studying}\quad
Figure \ref{fig:Tacc} shows the training accuracy of the student in the final epoch with and without the Self-Studying phase on the CIFAR-100 dataset. At very low-bit quantization, the model has low representative power and without SS, it gets stuck in a local minima as is reflected from the lower training accuracy. The SS phase helps the student start from a better initialization point and KD then guides it to a better local minima, hence increasing the training accuracy. Note that although training accuracy of QKD seems quite lower than others, its test accuracy is better than others, as was shown in Table \ref{table:cifar100}. This tendency is usually seen when using KD in general full-precision domain because KD regularizes the network for enhanced generalization performance. In QKD, the gap between blue and red one is large compared to BL, meaning that it suffers from regularization effect more.

\noindent \textbf{Adaptability of teacher network}\quad In Co-studying, we train the teacher network with $\mathcal{L}^{T}_{KD}$ (\ref{eq:LKDT}) with the goal of making it's distribution more adaptable to that of the low-precision student. In Figure \ref{fig:KL}, we plot the KL divergence between the teacher and student class distributions during QKD training and during SS+AP* training for one of the CIFAR-100 experiments. The KL-divergence during QKD is consistently lower than that of SS + AP*. This indicates that our QKD makes the teacher more adaptable to the low-precision student in terms of the similarity of their class distribution.
%as compared to the fixed teacher in SS+AP*.         

\noindent \textbf{Powerful teacher network}\quad
In addition to improving teacher's adaptability to the quantized student network, we observed that the teacher network's accuracy significantly increases during the Co-studying phase. This can be attributed to the regularizing effect and the knowledge that is being transferred from the student's posterior distribution. Figure \ref{fig:teacheracc} shows the variation of teacher accuracy with epochs for different settings during the CS + TU phase, compared to a fixed teacher used in AP*. The improvement in teacher accuracy was most in the case of W4A4 and the least with W2A2. The reason could be that knowledge transferred by a W2A2 quantized network is limited.

\vspace{1mm}

\noindent \textbf{Reasoning behind Tutoring phase}\quad From Figure \ref{fig:teacheracc}, we note that the teacher accuracy saturates towards the end of co-studying. This is due to the higher representation power of the teacher. Considering this, we freeze the teacher and turn into the tutoring phase. This helps tremendously in terms of training speed since now only the student is being trained. Interestingly, we also see performance gains by using SS + CS + TU (50 + 35 + 35 epochs) instead of just SS + CS (50 + 70 epochs). To verify this, we used SS + CS to train ResNet18 at 2-bit and 4-bit with ResNet34 as teacher. The performance of SS + CS was 0.2\% lower in case of W4A4 and 0.5\% lower with W2A2 compared to QKD.

\vspace{1mm}

\noindent \textbf{Activations vs. class posterior}\quad The works \cite{zagoruyko2016paying,kim2018paraphrasing,heo2019comprehensive,romero2014fitnets} have shown promising performance in the full-precision (FP) domain (teacher and student are both FP) by using offline activation (feature map) distillation. To compare distilling activations and the softmaxed posterior used in our QKD in terms of effectiveness on training a quantized network, we transfer the activations of the last layer of the teacher (full-precision) using L2 loss to the student (low-precision) similar to \cite{kim2018paraphrasing,romero2014fitnets}. We use a simple regressor used in \cite{romero2014fitnets,heo2019comprehensive}. We will call this baseline as activation distillation (AD). We use ResNet-56 as a teacher. Table \ref{table:ad} shows comparison between AD and QKD. AD has reasonable performance at 4bits but it loses power when we use lower bit-widths. The reason is that the activation distributions between low-precision and full-precision are quite different so it would not be a good guidance for the student. Hence, we verify that using posterior is more useful than activations in terms of training a low-precision network. 

\begin{table}[t]
    \small
	\caption{\small{CIFAR-100 performance}}
	\centering
	\begin{adjustbox}{width=0.8\linewidth}
	\begin{tabular}{p{1.6cm} |c| c c c}
        \toprule
        	{Network} & {Method} & {W2A2} & {W3A3} & {W4A4} \\
			
			\midrule
	           
			\multirow{4}{1.6cm}{ResNet-32 \\ } & AD & 62.1 & 70.2 & 72.2 \\
			& SS + AD   &63.2 & 70.4 & 72.3  \\
			& QKD  & \textbf{66.4} & \textbf{72.2} & \textbf{73.3}  \\
        \bottomrule
	\end{tabular}
	\end{adjustbox}
	\label{table:ad}
\end{table}

%%%%%%%%%%%%%%%%%%%%%%%%%%% supplementary details %%%%%%%%%%%%%%
%  \begin{figure}[t]
%  	\centering
%  	\includegraphics[width=0.9\linewidth]{dist1.png}
%  	\caption{Activation map plots for quantized and full-precision networks \jh{will be in the supplementary details}}
%  	\label{fig:Dist}
%  \end{figure}
%%%%%%%%%%%%%%%%%%%%%%%%%%%%%%%%%%%%%%%%%%%%%%%%%%%%%%%%

\section{Conclusion }
In this paper, we propose a Quantization-aware Knowledge Distillation method that can be effectively applied to very low-bit quantization. We propose a combination of self-studying, co-studying and tutoring methods to effectively combine model quantization and KD, wherein we provide a comprehensive ablation study of the impact of each of these methods on the quantized model accuracy. We show how self-studying is important in alleviating the regularization effect imposed by KD
%by demonstrating its effect on the student's training accuracy. Apart from enhancing the performance of the low-precision student model, we show that co-studying can also enhance the full-precision teacher's accuracy.
%which later helps during tutoring. 
and how co-studying enhances the quantization performance by making the teacher more adaptable and more powerful in terms of accuracy. Overall, with an extensive set of experiments, we show that QKD gives significant performance boost over our baseline quantization-only method and outperforms the existing state-of-the-art approaches. We demonstrate QKD's results on networks with both standard and depthwise separable convolutions and show that we can recover the full-precision accuracy at as low as W3A3 quantization of the ResNet architectures and W6A6 quantization of MobileNetV2.

{\small
\bibliographystyle{ieee_fullname}
\bibliography{egbib_cvpr}
}

\end{document}